\documentclass[11pt,a4paper]{article}
\usepackage[russian,main=english]{babel}
\usepackage[hyperref]{acl2020}
\usepackage{times}
\usepackage[utf8]{inputenc}
\usepackage{latexsym}
\usepackage{booktabs}
\usepackage{multicol}
\usepackage{graphicx}
\usepackage{amsmath}
\usepackage{linguex}
\usepackage{xcolor}
\usepackage{pgf}
\usepackage{collcell}

\usepackage{microtype}

\usepackage{listings}
\aclfinalcopy % Uncomment this line for the final submission
\def\aclpaperid{1849} %  Enter the acl Paper ID here

\newcommand{\ApplyGradientRed}[1]{%
    \textcolor{red}{#1}
}
\newcommand{\ApplyGradientBlue}[1]{%
    \textcolor{blue}{#1}
}

\newcommand\ul{\underline}
\newcommand\fr{\foreignlanguage{russian}}
\newcommand\bl{\ApplyGradientBlue}
\newcommand\red{\ApplyGradientRed}

\newcommand\blfootnote[1]{%
  \begingroup
  \renewcommand\thefootnote{}\footnote{#1}%
  \addtocounter{footnote}{-1}%
  \endgroup
}

\title{Cross-Linguistic Syntactic Evaluation of Word Prediction Models}

\author{Aaron Mueller\textnormal{$^1$}\qquad
        Garrett Nicolai\textnormal{$^{1\dagger}$}\qquad
        Panayiota Petrou-Zeniou\textnormal{$^2$}\\
        \textbf{Natalia Talmina\textnormal{$^2$}\qquad
        Tal Linzen\textnormal{$^{1,2}$}}\\
        $^1$Department of Computer Science\\
        $^2$Department of Cognitive Science\\
        Johns Hopkins University\\
  \texttt{\{amueller, gnicola2, ppetrou1, talmina, tal.linzen\}@jhu.edu}}

\date{}

\begin{document}
\maketitle
\begin{abstract}
A range of studies have concluded that neural word prediction models can distinguish grammatical from ungrammatical sentences with high accuracy. However, these studies are based primarily on monolingual evidence from English. To investigate how these models' ability to learn syntax varies by language, we introduce CLAMS (Cross-Linguistic Assessment of Models on Syntax), a syntactic evaluation suite for monolingual and multilingual models. CLAMS includes subject-verb agreement challenge sets for English, French, German, Hebrew and Russian, generated from grammars we develop. We use CLAMS to evaluate LSTM language models as well as monolingual and multilingual BERT. Across languages, monolingual LSTMs achieved high accuracy on dependencies without attractors, and generally poor accuracy on agreement across object relative clauses. On other constructions, agreement accuracy was generally higher in languages with richer morphology. Multilingual models generally underperformed monolingual models. Multilingual BERT showed high syntactic accuracy on English, but noticeable deficiencies in other languages.
\end{abstract}

\setlength{\Exlabelwidth}{0.25em}
\setlength{\SubExleftmargin}{1.35em}

\section{Introduction}
\setlength{\belowcaptionskip}{-2pt}
Neural networks can be trained to predict words from their context with much greater accuracy than the architectures used for this purpose in the past. This has been shown to be the case for both recurrent neural networks \citep{rnn_lm,lstm_lm,jozefowicz2016exploring} and non-recurrent attention-based models  \citep{bert,gpt2}.\blfootnote{$^\dagger$ Work done while at Johns Hopkins University. Now in the University of British Columbia's Linguistics Department.}

To gain a better understanding of these models' successes and failures, in particular in the domain of syntax, 
proposals have been made for testing the models on subsets of the test corpus where successful word prediction crucially depends on a correct analysis of the structure of the sentence \cite{dupouxlinzen16}. A paradigmatic example is subject-verb agreement. In many languages, including English, the verb often needs to agree in number (here, singular or plural) with the subject (asterisks represent ungrammatical word predictions):

\ex.The key to the cabinets \underline{is}/*\underline{are} next to the coins.
    
To correctly predict the form of the verb (underlined), the model needs to determine that the head of the subject of the sentence---an abstract, structurally defined notion---is the word \textit{key} rather than \textit{cabinets} or \textit{coins}. 

The approach of sampling challenging sentences from a test corpus has its limitations. Examples of relevant constructions may be difficult to find in the corpus, and naturally occurring sentences often contain statistical cues (confounds) that make it possible for the model to predict the correct form of the verb without an adequate syntactic analysis \cite{gulordavacolorless18}. To address these limitations, a growing number of studies have used constructed materials, which improve experimental control and coverage of syntactic constructions \cite{marvinlinzen18,wilcox18gap,futrell2019neural,warstadt2019blimp}.
    
Existing experimentally controlled data sets---in particular, those targeting subject-verb agreement---have largely been restricted to English. As such, we have a limited understanding of the effect of the cross-linguistic variability in neural networks' syntactic prediction abilities. In this paper, we introduce the Cross-Linguistic Assessment of Models on Syntax (CLAMS) data set, which extends the subject-verb agreement component of the \citet{marvinlinzen18} challenge set to French, German, Hebrew and Russian. By focusing on a single linguistic phenomenon in related languages,\footnote{English, French, German and Russian are all Indo-European languages, and (Modern) Hebrew syntax exhibits European areal influence (for different perspectives, see  \citealt{wexler1990schizoid,zuckermann2006new,zeldes2013modern}).} we can directly compare the models' performance across languages. We see the present effort as providing a core data set that can be expanded in future work to improve coverage to other languages and syntactic constructions. To this end, we release the code for a simple grammar engineering framework that facilitates the creation and generation of syntactic evaluation sets.\footnote{\url{https://github.com/aaronmueller/clams}}

We use CLAMS to test two hypotheses. First, we hypothesize that a multilingual model would show transfer across languages with similar syntactic constructions, which would lead to improved syntactic performance compared to monolingual models. In experiments on LSTM language models (LMs), we do not find support for this hypothesis; contrarily, accuracy was lower for the multilingual model than the monolingual ones. Second, we hypothesize that language models would be better able to learn hierarchical syntactic generalizations in morphologically complex languages (which provide frequent overt cues to syntactic structure) than in morphologically simpler languages \citep{gulordavacolorless18,Lorimor2008,mccoy18pos}. We test this using LSTM LMs we train, and find moderate support for this hypothesis. 

In addition to our analysis of LSTM LMs, we demonstrate the utility of CLAMS for testing pre-trained word prediction models. We evaluate multilingual BERT \cite{bert}, a bidirectional Transformer model trained on a multilingual corpus, and find that this model performs well on English, has mixed syntactic abilities in French and German, and performs poorly on Hebrew and Russian. Its syntactic performance in English was somewhat worse than that of monolingual English BERT, again suggesting that interference between languages offsets any potential syntactic transfer.

\section{Background and Previous Work}

\subsection{Word Prediction Models}

Language models (LMs) are statistical models that estimate the probability of sequences of words---or, equivalently, the probability of the next word of the sentence given the preceding ones. Currently, the most effective LMs are based on neural networks that are trained to predict the next word in a large corpus. Neural LMs are commonly based on LSTMs \citep{lstm,lstm_lm} or non-recurrent attention-based architectures (Transformers, \citealt{transformer}). The results of existing studies comparing the performance of the two architectures on grammatical evaluations are mixed \cite{tran18rnnovertrans,martylinzen19}, and the best reported syntactic performance on English grammatical evaluations comes from LMs trained with explicit syntactic supervision \cite{kuncoro18structure,kuncoro19}. We focus our experiments in the present study on LSTM-based models, but view CLAMS as a general tool for comparing LM architectures.

A generalized version of the word prediction paradigm, in which a bidirectional Transformer-based encoder is trained to predict one or more words in arbitrary locations in the sentence, has been shown to be an effective pre-training method in systems such as BERT \citep{bert}. While there are a number of variations on this architecture \citep{raffel2019exploring,gpt2}, we focus our evaluation on the pre-trained English BERT and multilingual BERT.

\subsection{Acceptability Judgments}
Human acceptability judgments have long been employed in linguistics to test the predictions of grammatical theories \citep{chomsky57ss,schutze96empirical}. There are a number of formulations of this task; we focus on the one in which a speaker is expected to judge a contrast between two minimally different sentences (a minimal pair). For instance, the following examples illustrate the contrast between grammatical and ungrammatical subject-verb agreement on the second verb in a coordination of short \ref{vp_coord_short} and long \ref{vp_coord_long} verb phrases; native speakers of English will generally agree that the first underlined verb is more acceptable than the second one in this context. 

\ex.\label{vp_both}\textit{Verb-phrase coordination}:\label{vp_coord}
    \a. The woman laughs and \underline{talks}/*\underline{talk}.\label{vp_coord_short}
    \b. My friends play tennis every week and then \underline{get}/*\underline{gets} ice cream.\label{vp_coord_long}

In computational linguistics, acceptability judgments have been used extensively to assess the grammatical abilities of LMs \citep{dupouxlinzen16,lau2017grammaticality}. For the minimal pair paradigm, this is done by determining whether the LM assigns a higher probability to the grammatical member of the minimal pair than to the ungrammatical member. This paradigm has been applied to a range of constructions, including subject-verb agreement \citep{marvinlinzen18,coordinationlevy19}, negative polarity item licensing \citep{marvinlinzen18,jumelet-hupkes-2018-npi}, filler-gap dependencies \citep{chowdhury2018rnn,wilcox18gap}, argument structure \citep{kann19structure}, and several others \cite{warstadt2019blimp}.

To the extent that the acceptability contrast relies on a single word in a particular location, as in~\ref{vp_both}, this approach can be extended to bidirectional word prediction systems such as BERT, even though they do not assign a probability to the sentence \citep{bertgoldberg19}. As we describe below, the current version of CLAMS only includes contrasts of this category.

An alternative use of acceptability judgments in NLP involves training an encoder to classify sentences into acceptable and unacceptable, as in the Corpus of Linguistic Acceptability (CoLA, \citealt{cola}). This approach requires supervised training on acceptable and unacceptable sentences; by contrast, the prediction approach we adopt can be used to evaluate any word prediction model without additional training.

\subsection{Grammatical Evaluation Beyond English}

Most of the work on grammatical evaluation of word prediction models has focused on English. However, there are a few exceptions, which we discuss in this section. To our knowledge, all of these studies have used sentences extracted from a corpus rather than a controlled challenge set, as we propose. \citet{gulordavacolorless18} extracted English, Italian, Hebrew, and Russian evaluation sentences from a treebank.  \citet{multiling_transfer} trained a multilingual LM on a concatenated French and Italian corpus, and tested whether grammatical abilities transfer across languages. \citet{basque_syntax} reported an in-depth analysis of LSTM LM performance on agreement prediction in Basque, and \citet{ravfogel2019studying} investigated the effect of different syntactic properties of a language on RNNs' agreement prediction accuracy by creating synthetic variants of English.
Finally, grammatical evaluation has been proposed for machine translation systems for languages such as German and French \citep{sennrich17grammatical,nmt_syntax}.

\section{Grammar Framework}
To construct our challenge sets, we use a lightweight grammar engineering framework that we term \textbf{attribute-varying grammars} (AVGs). This framework provides more flexibility than the hard-coded templates of \citet{marvinlinzen18} while avoiding the unbounded embedding depth of sentences generated from a recursive context-free grammar (CFG, \citealt{chomsky56cfg}).
This is done using \textit{templates}, which consist of \textit{preterminals} (which have \textit{attributes}) and \textit{terminals}. A \textit{vary statement} specifies which preterminal attributes are varied to generate ungrammatical sentences.

Templates define the structure of the sentences in the evaluation set. This is similar to the expansions of the \texttt{S} nonterminal in CFGs. Preterminals are similar to nonterminals in CFGs: they have a left-hand side which specifies the name of the preterminal and the preterminal's list of attributes, and a right-hand side which specifies all terminals to be generated by the preterminal. However, they are non-recursive and their right-hand sides may not contain other preterminals; rather, they must define a list of terminals to be generated. This is because we wish to generate all possible sentences given the template and preterminal definitions; if there existed any recursive preterminals, there would be an infinite number of possible sentences. All preterminals have an attribute list which is defined at the same time as the preterminal itself; this list is allowed to be empty. A terminal is a token or list of space-separated tokens.

The \texttt{vary} statement specifies a list of preterminals and associated attributes for each. Typically, we only wish to vary one preterminal per grammar such that each grammatical case is internally consistent with respect to which syntactic feature is varied. The following is a simple example of an attribute-varying grammar:

\begin{lstlisting}
       <@\textcolor{red}{vary:}@> <@\textcolor{blue}{V[]}@>
       <@\textcolor{red}{S[]}@>    -> je <@\textcolor{blue}{V[}\textcolor{orange}{1,s}\textcolor{blue}{]}@>
       <@\textcolor{blue}{V[}\textcolor{orange}{1,s}\textcolor{blue}{]}@> -> pense
       <@\textcolor{blue}{V[}\textcolor{orange}{2,s}\textcolor{blue}{]}@> -> penses
       <@\textcolor{blue}{V[}\textcolor{orange}{1,p}\textcolor{blue}{]}@> -> pensons
       <@\textcolor{blue}{V[}\textcolor{orange}{2,p}\textcolor{blue}{]}@> -> pensez
\end{lstlisting}
Preterminals are \textcolor{blue}{blue} and attributes are \textcolor{orange}{orange}. Here, the first statement is the \texttt{\textcolor{red}{vary}} statement. This is followed by a template, with the special \texttt{S} keyword in \textcolor{red}{red}. All remaining statements are preterminal definitions. All attributes are specified within brackets as comma-separated lists; these may be multiple characters and even multiple words long, so long as they do not contain commas. The output of this AVG is as follows (\textcolor{violet}{True} indicates that the sentence is grammatical):

\begin{lstlisting}
      <@\textcolor{violet}{True}@>     je pense
      <@\textcolor{violet}{False}@>    je penses
      <@\textcolor{violet}{False}@>    je pensons
      <@\textcolor{violet}{False}@>    je pensez
\end{lstlisting}

This particular grammar generates all possible verb forms because the attribute list for \texttt{V} in the \texttt{vary} statement is empty, which means that we may generate any \texttt{V} regardless of attributes. One may change which incorrect examples are generated by changing the \texttt{vary} statement; for example, if we change \texttt{V[]} to \texttt{V[1]}, we would only vary over verbs with the~\texttt{1} (first-person) attribute, thus generating \textit{je pense} and *\textit{je pensons}. One may also add multiple attributes within a single \texttt{vary} preterminal (implementing a logical AND) or multiple semicolon-separated \texttt{vary} preterminals (a logical OR). Changing \texttt{V[]} to \texttt{V[1,s]} in the example above would generate all first-person singular V terminals (here, \textit{je pense}). If instead we used \texttt{V[1]; V[s]}, this would generate all V terminals with either first-person and/or singular attributes (here, \textit{je pense}, *\textit{je penses}, and *\textit{je pensons}).

\section{Syntactic Constructions}

We construct grammars in French, German, Hebrew and Russian for a subset of the English constructions from \citet{marvinlinzen18}, shown in Figure~\ref{fig:examples}. These are implemented as AVGs by native or fluent speakers of the relevant languages who have academic training in linguistics.\footnote{The German grammar was created by a non-native speaker but was then validated by native speakers.}

A number of the constructions used by \citeauthor{marvinlinzen18} are English-specific. None of our languages besides English allow relative pronoun dropping, so we are unable to compare performance across languages on reduced relative clauses (\textit{the author the farmers like \underline{smile/*smiles}}). Likewise, we exclude \citeauthor{marvinlinzen18}'s sentential complement condition, which relies on the English-specific ability to omit complementizers (\textit{the bankers knew the officer \underline{smiles/*smile}}).

The \citet{marvinlinzen18} data set includes two additional structure-sensitive phenomena other than subject-verb agreement: reflexive anaphora and negative polarity item licensing. We do not include reflexive anaphora, as our languages vary significantly in how those are implemented. French and German, for example, do not distinguish singular from plural third-person reflexive pronouns. Similarly, negative polarity items (NPIs) have significantly different distributions across languages, and some of our evaluation languages do not even have items comparable to English NPIs \citep{npi08}.

\newlength{\vs}
\setlength{\vs}{0.5\baselineskip}
\begin{figure}
        \rule{\columnwidth}{1pt}
        \centering
        \resizebox{0.95\columnwidth}{!}{
        
        \begin{minipage}{\columnwidth}
            \raggedright
            \vspace{0.2cm}
         \textit{Simple Agreement}:\\
         The author \underline{laughs/*laugh}.

         \vspace{\vs}\textit{Across a Prepositional Phrase:}\\
         The farmer near the parents \underline{smiles/*smile}.

        \vspace{\vs}\textit{Across a Subject Relative Clause:}\\
        The officers that love the skater \underline{*smiles/smile}.

        \vspace{\vs}\textit{Short Verb Phrase Coordination:}\\
        The senator smiles and \underline{laughs/*laugh}.

        \vspace{\vs}\textit{Long Verb Phrase Coordination:}\\
The manager writes in a journal every day and \underline{likes/*like} to watch television shows.

        \vspace{\vs}\textit{Across Object Relative Clause:}\\
        The farmer that the parents love \underline{swims/*swim}.

        \vspace{\vs}\textit{Within Object Relative Clause:}\\
        The farmer that the parents \underline{*loves/love} swims.

        \vspace{0.2cm}
        \end{minipage}
    }

        \rule{\columnwidth}{1pt}
    \caption{Syntactic constructions used in CLAMS. Only English examples are shown; for examples in other languages, see Appendix A. Ungrammatical forms are marked with asterisks.}
    \label{fig:examples}
\end{figure}

We attempt to use translations of all terminals in \citet{marvinlinzen18}. In cases where this is not possible (due to differences in LM vocabulary across languages), we replace the word with another in-vocabulary item. See Appendix~\ref{sec:data_sizes} for more detail on vocabulary replacement procedures.

For replicability, we observe only third-person singular vs.\ plural distinctions (as opposed to all possible present-tense inflections) when replicating the evaluation sets of \citet{marvinlinzen18} in any language.

\section{Experimental Setup}

\subsection{Corpora}
Following \citet{gulordavacolorless18}, we download recent Wikipedia dumps for each of the languages, strip the Wikipedia markup using WikiExtractor,\footnote{\url{https://github.com/attardi/wikiextractor}} and use TreeTagger\footnote{\url{https://www.cis.uni-muenchen.de/~schmid/tools/TreeTagger/}} to tokenize the text and segment it into sentences. We eliminate sentences with more than 5\% unknown words.

Our evaluation is within-sentence rather than across sentences. Thus, to minimize the availability of cross-sentential dependencies in the training corpus, we shuffle the preprocessed Wikipedia sentences before extracting them into train/dev/test corpora. The corpus for each language consists of approximately 80 million tokens for training, as well as 10 million tokens each for development and testing. We generate language-specific vocabularies containing the 50,000 most common tokens in the training and development set; as is standard, out-of-vocabulary tokens in the training, development, and test sets are replaced with \texttt{<unk>}.

\subsection{Training and Evaluation}
We experiment with recurrent LMs and Transformer-based bidirectional encoders. LSTM LMs are trained for each language using the best hyperparameters in \citet{martylinzen19}.\footnote{\label{hyperparameters} Specifically, we use 2-layer word-level LSTMs with 800 hidden units in each layer, 800-dimensional word embeddings, initial learning rate $20.0$ (annealed after any epoch in which validation perplexity did not improve relative to the previous epoch), batch size $20$, and dropout probability $0.2$.} We will refer to these models as \textit{monolingual} LMs. We also train a \textit{multilingual} LSTM LM over all of our languages. The training set for this model is a concatenation of all of the individual languages' training corpora. The validation and test sets are concatenated in the same way, as are the vocabularies. We use the same hyperparameters as the monolingual models (Footnote~\ref{hyperparameters}). At each epoch, the corpora are randomly shuffled before batching; as such, each training batch consists with very high probability of sentences from multiple languages.

\begin{table*}[h!t]
    \centering
    \resizebox{\linewidth}{!}{
    \begin{tabular}{l  c c  c c  c c  c c  c c}
    \toprule
    & \multicolumn{2}{c}{English} & \multicolumn{2}{c}{French} & 
    \multicolumn{2}{c}{German} & 
    \multicolumn{2}{c}{Hebrew} & \multicolumn{2}{c}{Russian} \\
    \cmidrule(lr){2-3} \cmidrule(lr){4-5} \cmidrule(lr){6-7} \cmidrule(lr){8-9} \cmidrule(lr){10-11}
    & Mono & Multi & Mono & Multi & Mono & Multi & Mono & Multi & Mono & Multi \\\midrule
    \bf{Test Perplexity}& 57.90 & 66.13 & 35.48 & 57.40 & 46.31 & 61.06 & 48.78 & 61.85 & 35.09 & 54.61 \\\midrule\midrule
    Simple agreement & \bf{1.00} & \bf{1.00} & \bf{1.00} & \bf{1.00} & \bf{1.00} & 0.96 & 0.95 & 0.96 & 0.91 & 0.75 \\
    VP coordination (short) & 0.94 & 0.96 & 0.97 & 0.85 & 0.99 & \bf{1.00} & \bf{1.00} & 0.95 & 0.98 & 0.92 \\
    VP coordination (long) & 0.76 & 0.69 & 0.85 & 0.72 & \bf{0.96} & 0.73 & 0.84 & 0.70 & 0.86 & 0.72 \\
    Across subject rel.\ clause & 0.60 & 0.63 & 0.71 & 0.70 & \bf{0.94} & 0.74 & 0.91 & 0.84 & 0.88 & 0.86 \\
    Within object rel.\ clause & 0.89 & 0.79 & 0.99 & 0.99 & 0.74 & 0.69 & \bf{1.00} & 0.88 & 0.95 & 0.88 \\
    Across object rel.\ clause & 0.55 & 0.52 & 0.52 & 0.52 & \bf{0.81} & 0.74 & 0.56 & 0.54 & 0.60 & 0.57 \\
    Across prepositional phrase & 0.63 & 0.61 & 0.74 & 0.63 & \bf{0.89} & 0.82 & 0.88 & 0.82 & 0.76 & 0.61 \\
    \midrule
    \bf{Average accuracy} & 0.77 & 0.74 & 0.83 & 0.78 & \bf{0.90} & 0.81 & 0.88 & 0.81 & 0.85 & 0.76 \\
    \bottomrule
    \end{tabular}}
    \caption{LSTM LM test perplexities and accuracies on CLAMS across languages for the language-specific monolingual models and for our multilingual model. Results are averaged across five random initializations. Chance accuracy is $0.5$. Boldfaced numbers indicate the model that achieved the highest performance on a given construction \textit{across languages}.}
    \label{tab:lstm_lower_results}
\end{table*}

To obtain LSTM accuracies, we compute the total probability of each of the sentences in our challenge set, and then check within each minimal set whether the grammatical sentence has higher probability than the ungrammatical one. Because the syntactic performance of LSTM LMs has been found to vary across weight initializations \citep{mccoy18pos,kuncoro19}, we report mean accuracy over five random initializations for each LM. See Appendix~\ref{sec:variance} for standard deviations across runs on each test construction in each language.

We evaluate the syntactic abilities of multilingual BERT (mBERT, \citealt{bert}) using the approach of \citet{bertgoldberg19}. Specifically, we mask out the focus verb, obtain predictions for the masked position, and then compare the scores assigned to the grammatical and ungrammatical forms in the minimal set. We use the scripts provided by Goldberg\footnote{\url{https://github.com/yoavg/bert-syntax}} without modification, with the exception of using \texttt{bert-base-multilingual-cased} to obtain word probabilities. This approach is not equivalent to the method we use to evaluate LSTM LMs, as LSTM LMs score words based only on the left context, whereas BERT has access to left and right contexts. In some cases, mBERT's vocabulary does not include the focus verbs that we vary in a particular minimal set. In such cases, if either or both verbs were missing, we skip that minimal set and calculate accuracies without the sentences contained therein.

\section{Results}

\subsection{LSTMs}\label{sec:lstm_results}

\begin{figure}[t]
    \centering
    \includegraphics[width=1.0\columnwidth]{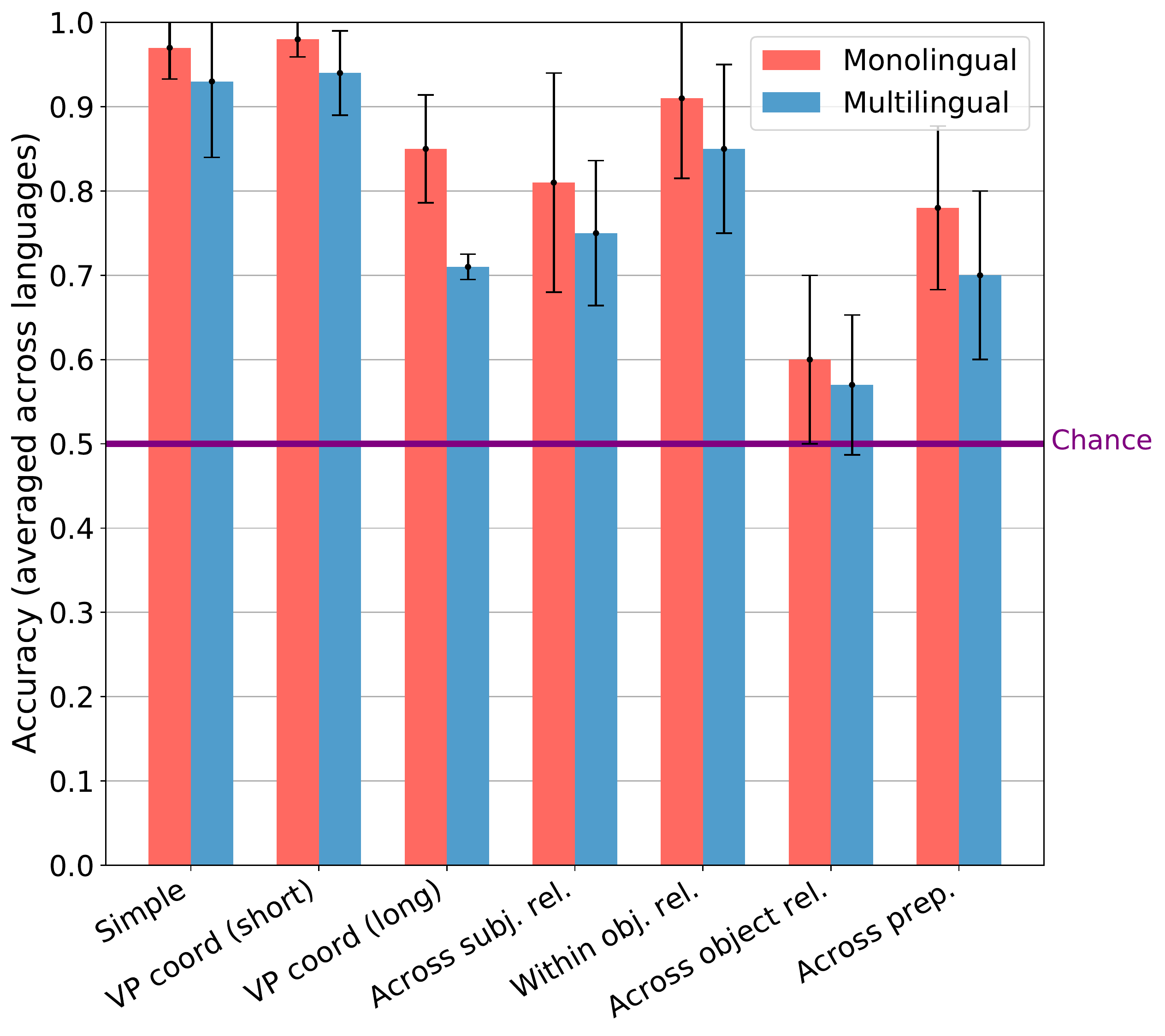}
    \caption{Mean accuracy (bars) and standard deviation (whiskers) for LSTM LMs over all languages for each stimulus type. \textbf{Note}: these are means over languages per-case, whereas the numbers in Table~\ref{tab:lstm_lower_results} are means over cases per-language.}
    \label{fig:mean_std}
\end{figure}

The overall syntactic performance of the monolingual LSTMs was fairly consistent across languages (Table~\ref{tab:lstm_lower_results} and Figure~\ref{fig:mean_std}). Accuracy on short dependencies without attractors---Simple Agreement and Short VP Coordination---was close to perfect in all languages. This suggests that all monolingual models learned the basic facts of agreement, and were able to apply them to the vocabulary items in our materials. At the other end of the spectrum, performance was only slightly higher than chance in the Across an Object Relative Clause condition for all languages except German, suggesting that LSTMs tend to struggle with center embedding---that is, when a subject-verb dependency is nested within another dependency of the same kind \cite{marvinlinzen18,noji2020analysis}.

There was higher variability across languages in the remaining three constructions. The German models had almost perfect accuracy in Long VP Coordination and Across Prepositional Phrase, compared to accuracies ranging between $0.76$ and $0.87$ for other languages in those constructions. The Hebrew, Russian, and German models showed very high performance on the Across Subject Relative Clause condition: $\geq 0.88$ compared to $0.6$--$0.71$ in other languages (recall that all our results are averaged over five runs, so this pattern is unlikely to be due to a single outlier).

With each of these trends, German seems to be a persistent outlier. This could be due to its marking of cases in separate article tokens---a unique feature among the languages evaluated here---or some facet of its word ordering or unique capitalization rules. In particular, subject relative clauses and object relative clauses have the same word order in German, but are differentiated by the case markings of the articles and relative pronouns. More investigation will be necessary to determine the sources of this deviation.

For most languages and constructions, the multilingual LM performed worse than the monolingual LMs, even though it was trained on five times as much data as each of the monolingual ones. Its average accuracy in each language was at least 3 percentage points lower than that of the corresponding monolingual LMs. Although all languages in our sample shared constructions such as prepositional phrases and relative clauses, there is no evidence that the multilingual LM acquired abstract representations that enable transfer across those languages; if anything, the languages interfered with each other. The absence of evidence for syntactic transfer across languages is consistent with the results of \citet{multiling_transfer2}, who likewise found no evidence of transfer in an LSTM LM trained on two closely related languages (French and Italian). One caveat is that the hyperparameters we chose for all of our LSTM LMs were based on a monolingual LM \cite{martylinzen19}; it is possible that the multilingual LM would have been more successful if we had optimized its hyperparameters separately (e.g., it might benefit from a larger hidden layer).

\begin{table*}[h]
    \centering
    \resizebox{0.75\textwidth}{!}{
    \begin{tabular}{l c c c c c}
    \toprule
    & English & French & German & Hebrew & Russian\\
    \midrule
    Simple agreement & \bf{1.00} & \bf{1.00} & 0.95 & 0.70 & 0.65\\
    VP coordination (short) & \bf{1.00} & \bf{1.00} & 0.97 & 0.91 & 0.80\\
    VP coordination (long) &  0.92 & 0.98 & \bf{1.00} & 0.73 & ---\\
    Across subject relative clause & \bf{0.88} & 0.57 & 0.73 & 0.61 & 0.70\\
    Within object relative clause & 0.83 & --- & --- & --- & ---\\
    Across object relative clause & 0.87 & 0.86 & \bf{0.93} & 0.55 & 0.67\\
    Across prepositional phrase & 0.92 & 0.57 & \bf{0.95} & 0.62 & 0.56\\
    \bottomrule
    \end{tabular}
    }
    \caption{Multilingual BERT accuracies on CLAMS. If a hyphen is present, this means that all focus verbs for that particular language and construction were out-of-vocabulary. Chance accuracy is $0.5$.}
    \label{tab:bert_multiling}
\end{table*}

These findings also suggest that test perplexity and subject-verb agreement accuracy in syntactically complex contexts are not strongly correlated cross-linguistically. This extends one of the results of \citet{kuncoro19}, who found that test perplexity and syntactic accuracy were not necessarily strongly correlated within English. Finally, the multilingual LM's perplexity was always higher than that of the monolingual LMs. At first glance, this contradicts the results of \citet{ostling17}, who observed lower perplexity in LMs trained on a small number of very similar languages (e.g., Danish, Swedish, and Norwegian) than in LMs trained on just one of those languages. However, their perplexity rose precipitously when trained on more languages and/or less-related languages---as we have here.

\subsection{BERT and mBERT}

Table~\ref{tab:bert_multiling} shows mBERT's accuracies on all stimuli.  Performance on CLAMS was fairly high in the languages that are written in Latin script (English, French and German). On English in particular, accuracy was high across conditions, ranging between $0.83$ and $0.88$ for sentences with relative clauses, and between $0.92$ and $1.00$ for the remaining conditions. Accuracy in German was also high: above $0.90$ on all constructions except Across Subject Relative Clause, where it was $0.73$. French accuracy was more variable: high for most conditions, but low for Across Subject Relative Clause and Across Prepositional Phrase.

In all Latin-script languages, accuracy on Across an Object Relative Clause was much higher than in our LSTMs. However, the results are not directly comparable, for two reasons. First, as we have mentioned, we followed \citet{bertgoldberg19} in excluding the examples whose focus verbs were not present in mBERT's vocabulary; this happened frequently (see Appendix~\ref{sec:data_sizes} for statistics). Perhaps more importantly, unlike the LSTM LMs, mBERT has access to the right context of the focus word; in Across Object Relative Clause sentences (\textit{the farmers that the lawyer likes \underline{smile/*smiles}.}), the period at the end of the sentence may indicate to a bidirectional model that the preceding word (\textit{smile/smiles}) is part of the main clause rather than the relative clause, and should therefore agree with \textit{farmers} rather than \textit{lawyer}.

In contrast to the languages written in Latin script, mBERT's accuracy was noticeably lower on Hebrew and Russian---even on the Simple Agreement cases, which do not pose any syntactic challenge. Multilingual BERT's surprisingly poor syntactic performance on these languages may arise from the fact that mBERT's vocabulary (of size 110,000) is shared across all languages, and that a large proportion of the training data is likely in Latin script. While \citet{bert} reweighted the training sets for each language to obtain a more even distribution across various languages during training, it remains the case that most of the largest Wikipedias are written in languages which use Latin script, whereas Hebrew script is used only by Hebrew, and the Cyrillic script, while used by several languages, is not as well-represented in the largest Wikipedias.

\begin{table}[h]
    \centering
    \resizebox{\columnwidth}{!}{
    \begin{tabular}{l r r}
    \toprule
    & Mono & Multi \\\midrule
    \multicolumn{3}{l}{\textsc{Subject-Verb Agreement}}\\
    Simple & 1.00 & 1.00 \\
    In a sentential complement & 0.83 & \bf{1.00} \\
    VP coordination (short) & 0.89 & \bf{1.00} \\
    VP coordination (long) & \bf{0.98} & 0.92 \\
    Across subject rel.\ clause & 0.84 & \bf{0.88} \\
    Within object rel.\ clause & \bf{0.95} & 0.83 \\
    Within object rel.\ clause (no \emph{that}) & \bf{0.79} & 0.61 \\
    Across object rel.\ clause & \bf{0.89} & 0.87 \\
    Across object rel.\ clause (no \emph{that}) & \bf{0.86} & 0.64 \\
    Across prepositional phrase & 0.85 & \bf{0.92} \\\midrule
    \textbf{Average accuracy} & \bf{0.89} & 0.87
    \\\midrule\midrule
    \multicolumn{3}{l}{\textsc{Reflexive Anaphora}}\\
    Simple & \bf{0.94} & 0.87 \\
    In a sentential complement & 0.89 & 0.89 \\
    Across a relative\ clause & \bf{0.80} & 0.74 \\\midrule
    \textbf{Average accuracy} & \bf{0.88} & 0.83 \\
    \bottomrule
    \end{tabular}}
    \caption{English BERT (base) and multilingual BERT accuracies on the English stimuli from \citet{marvinlinzen18}. Monolingual results are taken from \citet{bertgoldberg19}.}
    \label{tab:BERTvmBERT}
\end{table}

We next compare the performance of monolingual and multilingual BERT. Since this experiment is not limited to using constructions that appear in all of our languages, we use additional constructions from \citet{marvinlinzen18}, including reflexive anaphora and reduced relative clauses (i.e., relative clauses without \textit{that}). We exclude their negative polarity item examples, as the two members of a minimal pair in this construction differ in more than one word position.

The results of this experiment are shown in Table~\ref{tab:BERTvmBERT}. Multilingual BERT performed better than English BERT on Sentential Complements, Short VP Coordination, and Across a Prepositional Phrase, but worse on Within an Object Relative Clause, Across an Object Relative Clause (no relative pronoun), and in Reflexive Anaphora Across a Relative Clause. The omission of the relative pronoun \textit{that} caused a sharp drop in performance in mBERT, and a milder drop in English BERT. Otherwise, both models had similar accuracies on other stimuli. These results reinforce the finding in LSTMs that multilingual models generally underperform monolingual models of the same architecture, though there are specific contexts in which they can perform slightly better.

\subsection{Morphological Complexity vs.\ Accuracy}

Languages vary in the extent to which they indicate the syntactic role of a word using overt morphemes. In Russian, for example, the subject is generally marked with a suffix indicating nominative case, and the direct object with a different suffix indicating accusative case. Such case distinctions are rarely indicated in English, with the exception of pronouns (\textit{he} vs. \textit{him}). English also displays significant syncretism: morphological distinctions that are made in some contexts (e.g., \textit{eat} for plural subjects vs. \textit{eats} for singular subjects) are neutralized in others (\textit{ate} for both singular and plural subjects). We predict that greater morphological complexity, which is likely to correlate with less syncretism, will provide more explicit cues to hierarchical syntactic structure,\footnote{For more evidence that explicit cues to structural information can aid syntactic performance, see Appendix~\ref{sec:capitalization_matters}.} and thus result in increased overall accuracy on a given language.

To measure the morphological complexity of a language, we use the C$_{\text{WALS}}$ metric of \citet{bentz16morph}: $\frac{\sum_{i=1}^n f_i}{n}$.
This is a typological measure of complexity based on the World Atlas of Language Structures (WALS, \citealt{wals}), where $f_i$ refers to a morphological feature value normalized to the range $[0,1]$.\footnote{For example, if WALS states that a language has negative morphemes, $f_{28}$ is $1$; otherwise, $f_{28}$ is $0$.} This essentially amounts to a mean over normalized values of quantified morphological features. Here, $n$ is 27 or 28 depending on the number of morphological categorizations present for a given language in WALS.

\begin{figure}[th]
    \centering
    \includegraphics[width=\columnwidth]{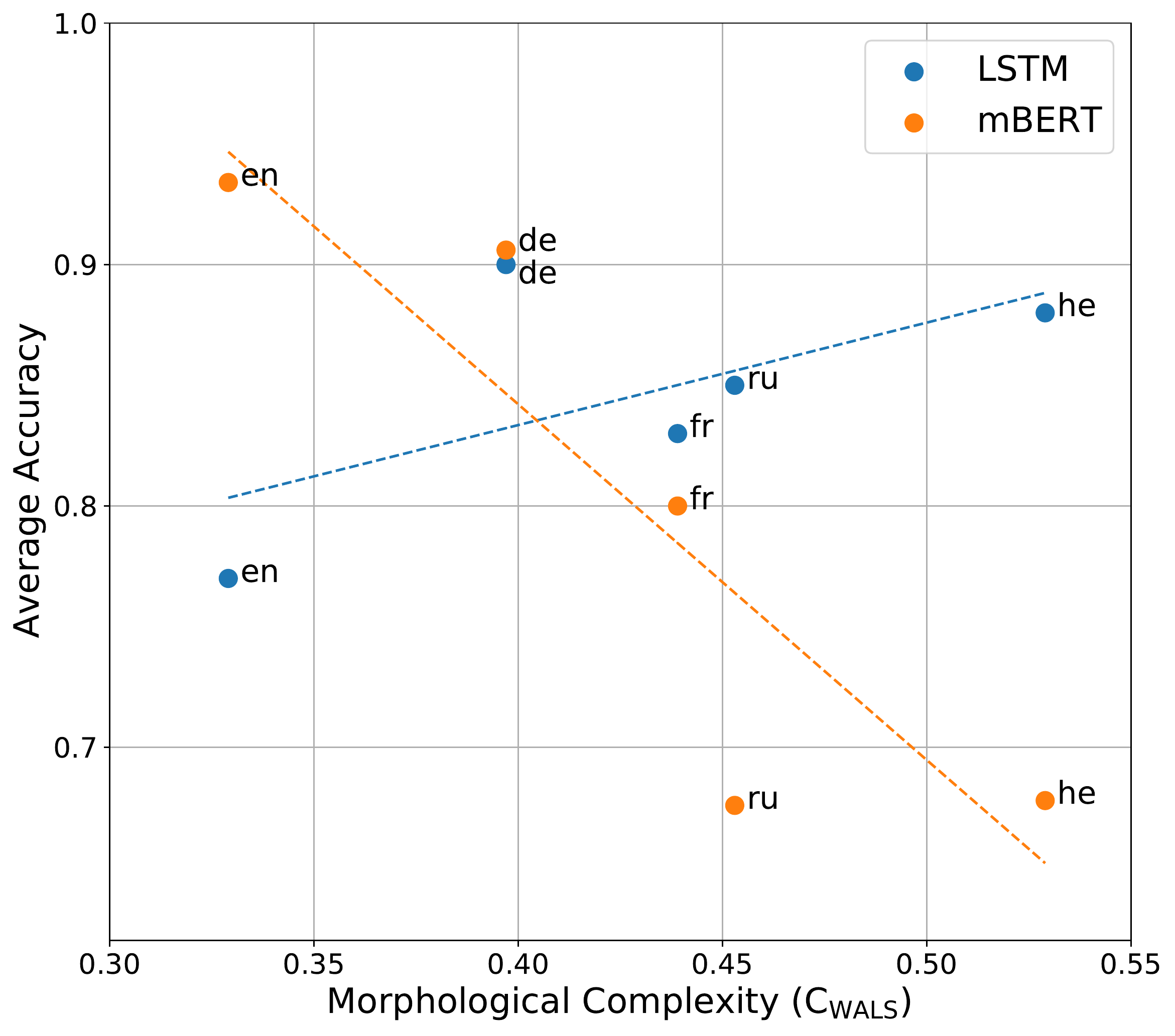}
    \caption{Morphological complexities against average accuracies per-language for LSTMs and mBERT.}
    \label{fig:morph_acc}
\end{figure}

Does the morphological complexity of a language correlate with the syntactic prediction accuracy of LMs trained on that language? In the LSTM LMs (Table~\ref{tab:lstm_lower_results}), the answer is generally yes, but not consistently. We see higher average accuracies for French than English (French has more distinct person/number verb inflections), higher for Russian than French, and higher for Hebrew than Russian (Hebrew verbs are inflected for person, number, \emph{and} gender). However, German is again an outlier: despite its notably lower complexity than Hebrew and Russian, it achieved a higher average accuracy. The same reasoning applied in Section~\ref{sec:lstm_results} for German's deviation from otherwise consistent trends applies to this analysis as well.

Nonetheless, the Spearman correlation between morphological complexity and average accuracy including German is~$0.4$; excluding German, it is~$1.0$. Because we have the same amount of training data per-language in the same domain, this could point to the importance of having explicit cues to linguistic structure such that models can learn that structure. While more language varieties need to be evaluated to determine whether this trend is robust, we note that this finding is consistent with that of \citet{ravfogel2019studying}, who compared English to a synthetic variety of English augmented with case markers and found that the addition of case markers increased LSTM agreement prediction accuracy.

We see the opposite trend for mBERT (Table~\ref{tab:bert_multiling}): if we take the average accuracy over all stimulus types for which we have scores for all languages---i.e., all stimulus types except Long VP Coordination and Within an Object Relative Clause---then we see a correlation of $\rho=-0.9$. In other words, accuracy is likely to \emph{decrease} with increasing morphological complexity. This unexpected inverse correlation may be an artifact of mBERT's limited vocabulary, especially in non-Latin scripts. Morphologically complex languages have more unique word types. In some languages, this issue can be mitigated to some extent by splitting the word into subword units, as BERT does; however, the effectiveness of such a strategy would be limited at best in a language with non-concatenative morphology such as Hebrew. Finally, we stress that the exclusion of certain stimulus types and the differing amount of training data per-language act as confounding variables, rendering a comparison between mBERT and LSTMs difficult.

\section{Conclusions}
In this work, we have introduced the CLAMS data set for cross-linguistic syntactic evaluation of word prediction models, and used it to to evaluate monolingual and multilingual versions of LSTMs and BERT. The design conditions of \citet{marvinlinzen18} and our cross-linguistic replications rule out the possibility of memorizing the training data or relying on statistical correlations/token collocations. Thus, our findings indicate that LSTM language models can distinguish grammatical from ungrammatical subject-verb agreement dependencies with considerable overall accuracy across languages, but their accuracy declines on some constructions (in particular, center-embedded clauses). We also find that multilingual neural LMs in their current form do not show signs of transfer across languages, but rather harmful interference. This issue could be mitigated in the future with architectural changes to neural LMs (such as better handling of morphology), more principled combinations of languages (as in \citealt{multiling_transfer2}), or through explicit separation between languages during training (e.g., using explicit language IDs).

Our experiments on BERT and mBERT suggest (1) that mBERT shows signs of learning syntactic generalizations in multiple languages, (2) that it learns these generalizations better in some languages than others, and (3) that its sensitivity to syntax is lower than that of monolingual BERT. It is possible that its performance drop in Hebrew and Russian could be mitigated with fine-tuning on more data in these languages.

When evaluating the effect of the morphological complexity of a language on the LMs' syntactic prediction accuracy, we found that recurrent neural LMs demonstrate better hierarchical syntactic knowledge in morphologically richer languages. Conversely, mBERT demonstrated moderately better syntactic knowledge in morphologically \emph{simpler} languages. Since CLAMS currently includes only five languages, this correlation should be taken as very preliminary. In future work, we intend to expand the coverage of CLAMS by incorporating language-specific and non-binary phenomena (e.g., French subjunctive vs.\ indicative and different person/number combinations, respectively), and by expanding the typological diversity of our languages.

\section*{Acknowledgments}

This material is based on work supported by the National Science Foundation Graduate Research Fellowship Program under Grant No.\ 1746891. Any opinions, findings, and conclusions or recommendations expressed in this material are those of the authors and do not necessarily reflect the views of the National Science Foundation or the other supporting agencies. Additionally, this work was supported by a Google Faculty Research Award to Tal Linzen, and by the United States--Israel Binational Science Foundation (award 2018284).

\bibliography{clams_acl}
\bibliographystyle{acl_natbib}

% APPENDIX

\setlength{\Exlabelwidth}{0.25em}
\setlength{\SubExleftmargin}{1.35em}

\appendix

\section{Linguistic Examples}
This section provides examples of the syntactic structures included in the CLAMS dataset across languages. For Hebrew, we transliterate its original right-to-left script into the left-to-right Latin script; this makes labeling and glossing more consistent across languages. Hebrew was \textbf{not} transliterated in the training/development/test corpora or in the evaluation sets. In all examples, (a) is English, (b)~is French, (c) is German, (d) is Hebrew, and (e) is Russian.

The first case is simple agreement. This simply involves agreeing a verb with its adjacent subject, which should pose little challenge for any good language model regardless of syntactic knowledge.

\ex.\textit{Simple Agreement}:\label{ex:agree}
    \a. The surgeons \underline{laugh}/*\underline{laughs}.\label{ex:agree_en}
    \bg. Le pilote \underline{parle} / *\underline{parlent}.\label{ex:agree_fr}\\
         The pilot laughs / *laugh.\\
    \cg. Der Schriftsteller \ul{spricht} / *\ul{sprechen}.\label{ex:agree_de}\\
         The writer speaks / *speak.\\
    \dg. Ha meltsar \underline{yashen} / \underline{yeshenim}.\\
        The server sleeps / *sleep.\\
    \eg. \fr{Врачи} \fr{\underline{говорят}} / *\fr{\underline{говорит}}.\\
        Doctors speak / *speaks.\\

Short verb-phrase coordination introduces some slight distance between the subject and verb, though the presence of the previous verb should give a model a clue as to which inflection should be more probable.
\ex.\textit{VP coordination (short)}:\label{ex:vp_short}
    \a. The author swims and \underline{smiles}/*\underline{smile}.\label{ex:vp_short_en}
    \bg. Les directeurs parlent et \underline{déménagent} / *\underline{déménage}.\label{ex:vp_short_fr}\\
         The directors talk and move / *moves.\\
    \cg. Der Polizist schwimmt und \ul{lacht} / *\ul{lachen}.\\
         The police.officer swims and laughs / *laugh.\\
    \dg. Ha tabaxim rokdim ve \underline{soxim} / *\underline{soxe}.\\
    The cooks dance and swim / *swims.\\
    \eg. \fr{Профессор} \fr{старый} \fr{и} \fr{\ul{читает}} / *\fr{\ul{читают}}.\\
       Professor is.old and reads / *read.\\
         
Long verb-phrase coordination is similar, but makes each verb phrase much longer to introduce more distance and attractors between the subject and target verb.
\ex.\textit{VP coordination (long)}:\label{ex:vp_long}
    \a. The teacher knows many different foreign languages and \underline{likes}/*\underline{like} to watch television shows.\label{ex:vp_long_en}
    \bg. L' agriculteur écrit dans un journal tous les jours et \underline{préfère} / *\underline{préfèrent} jouer au tennis avec des collègues.\label{ex:vp_long_fr}\\
         The farmer writes in a journal all the days and prefers / *prefer to.play at.the tennis with some colleagues.\\
    \cg. Die Bauern sprechen viele verschiedene Sprachen und \underline{sehen} / *\underline{sieht} gern Fernsehprogramme.\\
     The farmers speak many various languages and watch / *watches gladly TV.shows.\\
    \dg. Ha tabax ohev litspot be toxniot televizya ve \underline{gar} / *\underline{garim} be merkaz ha ir.\\
    The cook likes to.watch in shows TV and lives / *live in center the city.\\
    \eg. \fr{Автор} \fr{знает} \fr{много} \fr{иностранных} \fr{языков} \fr{и} \fr{\ul{любит}} / *\fr{\ul{любят}} \fr{смотреть} \fr{телепередачи}.\\
     Author knows many foreign languages and likes / *like to.watch TV.shows.\\

Now we have more complex structures that require some form of structural knowledge if a model is to obtain the correct predictions with more than random-chance accuracy. Agreement across a subject relative clause involves a subject with an attached relative clause containing a verb and object, followed by the main verb. Here, the attractor is the object in the relative clause. (An attractor is an intervening noun between a noun and its associated finite verb which might influence a human's or model's decision as to which inflection to choose. This might be of the same person and number, or, in more difficult cases, a different person and/or number. It does not necessarily need to occur between the noun and its associated verb, though this does tend to render this task more difficult.)

\ex.\textit{Across a subject relative clause}:\label{ex:subj_rel}
    \a. The officers that love the chef \underline{are}/*\underline{is} old.\label{ex:subj_rel_en}
    \bg. Les chirurgiens qui détestent le garde \underline{retournent} / *\underline{retourne}. \label{ex:subj_rel_fr}\\
         The surgeons that hate the guard return / *returns\\
    \cg. Der Kunde, der die Architekten hasst, \ul{ist} / *\ul{sind} klein.\\
    The customer that the architects hates is / *are short.\\
    \dg. Ha menahel she ma'arits et ha shomer \underline{rats} / *\underline{ratsim}.\\
    The manager who admires \textsc{acc} the guard runs / *\underline{run}.\\
    \eg. \fr{Пилоты}, \fr{которые} \fr{понимают} \fr{агентов}, \fr{\ul{говорят}} / *\fr{\ul{говорит}}.\\
    Pilots that understand agents speak / *speaks.\\
         
Agreement within an object relative clause requires the model to inflect the proper verb inside of an object relative clause; the object relative clause contains a noun and an associated transitive verb whose object requirement is filled by the relative pronoun. The model must choose the proper verb inflection given the noun within the relative clause as opposed to the noun outside of it. This may seem similar to simple agreement, but we now have an attractor which appears before the noun of the target verb.
\ex.\textit{Within an object relative clause}:\label{ex:obj_rel_within}
    \a. The senator that the executives \underline{love}/*\underline{loves} laughs.\label{ex:obj_rel_within_en}
    \bg. Les professeurs que le chef \underline{admire} / *\underline{admirent} parlent. \label{ex:obj_rel_within_fr}\\
         The professors that the boss admires / *admire talk.\\
    \cg. Die Polizisten, die der Bruder \ul{hasst}, / *\ul{hassen}, sind alt\\
     The police.officers that the brother hates / *hate are old.\\
    \dg. Ha menahel she ha nahag \underline{ma'aritz} / *\underline{ma'aritsim} soxe.\\
    The manager that the driver admires / *admire swims.\\
    \eg. \fr{Сенаторы}, \fr{которых} \fr{рабочие} \fr{\ul{ищут}}, / *\fr{\ul{ищет}}, \fr{ждали}. \\
      Senators that workers seek / *seeks wait.\\

Agreement across an object relative clause is similar, but now the model must choose the correct inflection for the noun outside of the relative clause. This requires the model to capture long-range dependencies, and requires it to have the proper structural understanding to ignore the relative clause when choosing the proper inflection for the focus verb.
\ex.\textit{Across an object relative clause}:\label{ex:obj_rel_across}
    \a. The senator that the executives love \underline{laughs}/*\underline{laugh}.\label{ex:obj_rel_across_en}
    \bg. Les professeurs que le chef admire \underline{parlent} / *\underline{parle}.\label{ex:obj_rel_across_fr}\\
         The professors that the boss admires talk / *talks.\\
    \cg. Der Senator, den die Tänzer mögen, \ul{spricht} / *\ul{sprechen}.\\
    The senator that the dancers like speaks / *speak.\\
    \dg. Ha katsin she ha zamar ohev \ul{soxe} / *\ul{soxim}.\\
    The officer that the singer likes \ul{swims} / *\ul{swim}.\\
    \eg. \fr{Фермеры}, \fr{которых} \fr{танцоры} \fr{хотят}, \fr{\ul{большие}} / *\fr{\ul{большой}}.\\
      Farmers that dancers want are.big / *is.big. \\
         
Finally, agreement across a prepositional phrase entails placing a prepositional phrase after the subject; the prepositional phrase contains an attractor, which makes choosing the correct inflection more difficult.
\ex.\textit{Across a prepositional phrase}:\label{ex:prep}
    \a. The consultants behind the executive \underline{smile}/*\underline{smiles}.\label{ex:prep_en}
    \bg. Les clients devant l' adjoint \underline{sont} / *\underline{est} vieux.\label{ex:prep_fr}\\
         The clients in.front.of the deputy are / *is old.\\
    \cg. Der Lehrer neben den Ministern \ul{lacht} / *\ul{lachen}.\\
    The teacher next.to the ministers laughs / *laugh.\\
    \dg. Ha meltsarim leyad ha zamarim \underline{nos'im} / *\underline{nose'a}.\\
    The servers near the singers drive / *drives.\\
    \eg. \fr{Режиссёры} \fr{перед} \fr{агентами} \fr{\ul{маленькие}} / *\fr{\ul{маленький}}.\\
      Directors in.front.of agents are.small / *is.small. \\

\begin{table*}[t]
    \centering
    \resizebox{\linewidth}{!}{
    \begin{tabular}{l  c c  c c  c c  c c}
    \toprule
    & \multicolumn{2}{c}{English} & \multicolumn{2}{c}{French} & 
    \multicolumn{2}{c}{German} & \multicolumn{2}{c}{Russian} \\
    \cmidrule(lr){2-3} \cmidrule(lr){4-5} \cmidrule(lr){6-7} \cmidrule(lr){8-9}
    & Mono & Multi & Mono & Multi & Mono & Multi & Mono & Multi \\\midrule
    Simple agreement & --- & \red{-.02} & --- & \red{-.01} & --- & \bl{+.02} & \bl{+.02} & --- \\
    VP coordination (short) & \red{-.01} & --- & \bl{+.01} & \bl{+.14} & \red{-.02} & \red{-.01} & \red{-.03} & \red{-.01} \\
    VP coordination (long) & \red{-.03} & \bl{+.01} & \bl{+.04} & \red{-.02} & \red{-.06} & \bl{+.07} & \bl{+.04} & \bl{+.02} \\
    Across subject rel.\ clause & \bl{+.24} & \bl{+.07} & \bl{+.23} & \bl{+.15} & \red{-.03} & \bl{+.13} & \bl{+.02} & \bl{+.01} \\
    Within object rel.\ clause & --- & \red{-.04} & --- & \red{-.07} & --- & \red{-.02} & - & \red{-.03} \\
    Across object rel.\ clause & \bl{+.09} & \bl{+.02} & \bl{+.05} & \bl{+.03} & \bl{+.01} & \bl{+.09} & \bl{+.01} & - \\
    Across prepositional phrase & \bl{+.18} & \bl{+.11} & \bl{+.20} & \bl{+.20} & \bl{+.03} & \bl{+.03} & \bl{+.03} & \bl{+.02} \\
    \midrule
    \bf{Average accuracy} & \bl{+.06} & \bl{+.03} & \bl{+.07} & \bl{+.05} & \red{-.01} & \bl{+.05} & \bl{+.01} & \bl{+.03} \\
    \bottomrule
    \end{tabular}}
    \caption{Gains (positive, \textcolor{blue}{blue}) and losses (negative, \textcolor{red}{red}) in LSTM LM accuracies on CLAMS after capitalizing the first character of each evaluation example. Differences are relative to the results in Table~\ref{tab:lstm_lower_results}. Results are averaged across five random initializations.}
    \label{tab:lstm_capital_results}
\end{table*}

Some of the constructions used by \citet{marvinlinzen18} could not be replicated across languages. This includes reflexive anaphora, where none of our non-English languages use quite the same syntactic structures as English (or even to each other) when employing reflexive verbs and pronouns. Some do not even have separate reflexive pronouns for third-person singular and plural distinctions (like French and German). Moreover, the English reflexive examples rely on the syncretism between past-tense verbs for any English person and number,\footnote{For example, regardless of whether the subject is singular, plural, first- or third-person, etc., the past-tense of \emph{see} is always \emph{saw}.} whereas other languages often have different surface forms for different person and number combinations in the past tense. This would give the model a large clue as to which reflexive is correct. Thus, any results on reflexive anaphora would not be comparable cross-linguistically. See example~\ref{ex:reflexives} below for English, French, and German examples of the differences in reflexive syntax.

\ex.\textit{\ \ Reflexive anaphora across relative clause}:\label{ex:reflexives}
    \a. The author that the guards like injured \underline{himself}/*\underline{themselves}.\label{ex:reflexives_en}
    \bg. L' auteur que les gardes aiment \underline{s'} est blessé / *\underline{se} sont blessés.\label{ex:reflexives_fr}\\
         The author that the guards like \textsc{refl.3} has.\textsc{3s} injured.\textsc{s.masc} / \textsc{refl.3} have.\textsc{3p} injured.\textsc{p.masc}\\
    \cg. Der Autor, den die Wächter mögen, verletzte \underline{sich} / *verletzten \underline{sich}.\label{ex:reflexives_de}\\
         The author that the guards like injured.\textsc{3s} \textsc{refl.3} / injured.\textsc{3p} \textsc{refl.3}\\

\section{The Importance of Capitalization}\label{sec:capitalization_matters}

\begin{table*}[h]
    \centering
    \resizebox{\linewidth}{!}{
    \begin{tabular}{l  c c  c c  c c  c c  c c}
    \toprule
    & \multicolumn{2}{c}{English} & \multicolumn{2}{c}{French} & 
    \multicolumn{2}{c}{German} & 
    \multicolumn{2}{c}{Hebrew} & \multicolumn{2}{c}{Russian} \\
    \cmidrule(lr){2-3} \cmidrule(lr){4-5} \cmidrule(lr){6-7} \cmidrule(lr){8-9} \cmidrule(lr){10-11}
    & Mono & Multi & Mono & Multi & Mono & Multi & Mono & Multi & Mono & Multi \\\midrule
    Simple agreement & .00 & .00 & .00 & .00 & .00 & .02 & .01 & .01 & .01 & .07 \\
    VP coordination (short) & .01 & .00 & .01 & .05 & .02 & .00 & .01 & .01 & .02 & .02 \\
    VP coordination (long) & .06 & .08 & .05 & .09 & .04 & .07 & .06 & .06 & .04 & .06 \\
    Across subject rel.\ clause & .06 & .02 & .05 & .05 & .04 & .07 & .03 & .03 & .03 & .04 \\
    Within object rel.\ clause & .01 & .02 & .01 & .01 & .03 & .04 & .01 & .03 & .04 & .02 \\
    Across object rel.\ clause & .05 & .02 & .01 & .01 & .09 & .06 & .01 & .01 & .03 & .02 \\ % anim
    Across prepositional phrase & .02 & .02 & .02 & .02 & .06 & .03 & .03 & .04 & .02 & .01 \\ % anim
    \bottomrule
    \end{tabular}}
    \caption{Standard deviation of LSTM LM performance across five random weight initializations for all languages and stimulus types.}
    \label{tab:variance}
\end{table*}

As discovered in \citet{hao2020attribution}, capitalizing the first character of each test example improves the performance of language models in distinguishing grammatical from ungrammatical sentences in English. To test whether this finding holds cross-linguistically, we capitalize the first character of each of our test examples in all applicable languages. Hebrew has no capital-/lower-case distinction, so it is excluded from this analysis.

Table~\ref{tab:lstm_capital_results} contains the results and relative gains or losses of our LSTM language models on the capitalized stimuli compared to the lowercase ones. For all languages except German, we see a notable increase in the syntactic ability of our models. For German, we see a small drop in overall performance, but its performance was already exceptionally high in the lowercase examples (perhaps due to its mandatory capitalization of all nouns).

An interesting change is that morphological complexity no longer correlates with the overall syntactic performance across languages ($\rho=0.2$). Perhaps the capitalization acts as an explicit cue to syntactic structure by delineating the beginning of a sentence, thus supplanting the role of morphological cues in aiding the model to distinguish grammatical sentences.

Overall, it seems quite beneficial to capitalize one's test sentences before feeding them to a language model if one wishes to improve syntactic accuracy. The explanation given by \citet{hao2020attribution} is that \emph{The} essentially only appears sentence-initially, thus giving the model clues as to which noun (typically the token following \emph{The}) is the subject. Conversely, \emph{the} has a more varied distribution, as it may appear before essentially any noun in subject or object position; thus, it gives the model fewer cues as to which noun agrees with a given verb. This would explain the larger score increase for English and French (which employ articles in a similar fashion in CLAMS), as well as the milder increase for Russian (which does not have articles). However, it does not explain the decrease in performance on German. A deeper investigation of this trend per-language could reveal interesting trends about the heuristics employed by language models when scoring syntactically complex sentences.

\section{Performance Variance}\label{sec:variance}

\begin{table*}[h!]
    \centering
    %\resizebox{\linewidth}{!}{
    \begin{tabular}{l  c c c c c}
    \toprule
    & English & French & German & Hebrew & Russian \\\midrule
    Simple agreement & 140 & 280 & 140 & 140 & 280 \\
    VP coordination (short) & 840 & 980 & 980 & 980 & 980 \\
    VP coordination (long) & 400 & 500 & 500 & 500 & 500\\
    Across subject rel.\ clause & 11200 & 11200 & 11200 & 11200 & 10080 \\
    Within object rel.\ clause & 11200 & 11200 & 11200 & 11200 & 11200 \\
    Across object rel.\ clause & 11200 & 11200 & 11200 & 11200 & 11200\\ % anim
    %Across object rel.\ clause (inanim) & 0.63 & & 0.54 & & 0.56 & & - & & & \\
    Across prepositional phrase & 16800 & 14000 & 12600 & 5600 & 5880 \\ % anim
    %Across prepositional phrase (inanim) & 0.66 & & 0.88 & & 0.94 & & - & & & \\
    \bottomrule
    \end{tabular}
    \caption{Number of minimal sets for all languages and stimulus types using animate nouns.}
    \label{tab:examples_lstm}
\end{table*}

\begin{table*}[h]
    \centering
    \begin{tabular}{l r r r r r r}
    \toprule
    & \multicolumn{2}{c}{English} & & & & \\
    \cmidrule(lr){2-3}
    & Mono & Multi & French & German & Hebrew & Russian\\
    \midrule
    \multicolumn{7}{l}{\textsc{Subject-Verb Agreement}} \\
    Simple agreement & 120 & 80 & 40 & 100 & 20 & 80 \\
    In a sentential complement & 1440 & 960 & - & - & - & - \\
    VP coordination (short) & 720 & 480 & 140 & 700 & 140 & 280 \\
    VP coordination (long) & 400 & 240 & 100 & 300 & 100 & 0 \\
    Across subject rel.\ clause & 9600 & 6400 & 1600 & 5406 & 1600 & 2880 \\
    Within object rel.\ clause & 15960 & 5320 & 0 & 0 & 0 & 0 \\
    Within object rel.\ clause (no \emph{that}) & 15960 & 5320 & - & - & - & - \\
    Across object rel.\ clause & 19680 & 16480 & 1600 & 5620 & 1600 & 3200 \\
    Across object rel.\ clause (no \emph{that}) & 19680 & 16480 & - & - & - & -\\
    Across prepositional phrase & 19440 & 14640 & 2000 & 9000 & 800 & 1680 \\
    \midrule
    \multicolumn{7}{l}{\textsc{Reflexive Anaphora}} \\
    Simple & 280 & 280 & - & - & - & - \\
    In a sentential complement & 3360 & 3360 & - & - & - & - \\
    Across a rel.\ clause & 22400 & 22400 & - & - & - & - \\
    \bottomrule
    \end{tabular}
    \caption{Number of minimal sets used by BERT (English monolingual only) and mBERT for evaluation. The number of monolingual English examples is the same as in \citet{bertgoldberg19}. Hyphens indicate non-replicable stimulus types, and $0$ indicates that all focus verbs for a given stimulus type were out-of-vocabulary.}
    \label{tab:examples_bert}
\end{table*}

Previous work has found the variance of LSTM performance in syntactic agreement to be quite high \citep{mccoy18pos,kuncoro19}. In Table~\ref{tab:variance}, we provide the standard deviation of accuracy over five random initializations on all CLAMS languages and stimulus types. This value never exceeds 0.1, and tends to only exceed 0.05 in more difficult syntactic contexts.

For syntactic contexts without attractors, the standard deviation is generally low. In more difficult cases like Across a Subject Relative Clause and Long VP Coordination, we see far higher variance. In Across an Object Relative Clause, however, the standard deviation is quite low despite this being the case on which language models struggled most; this is likely due to the consistently at-chance performance on this case, further showcasing the difficulty of learning syntactic agreements in such contexts.

On cases where German tended to deviate from the general trends seen in other languages, we see our highest standard deviations. Notably, the performance of German LMs in Across an Object Relative Clause and Across a Prepositional Phrase varies far more than other languages for the same stimulus type.

\section{Evaluation Set Sizes}\label{sec:data_sizes}

Here, we describe the size of the various evaluation set replications. These will differ for the LSTMs, BERT, and mBERT, as the two latter models sometimes do not contain the varied focus verb for a particular minimal set.

Table~\ref{tab:examples_lstm} displays the number of minimal sets per language and stimulus type (with animate nouns only) in our evaluation sets; the total number of sentences (grammatical \emph{and} ungrammatical) is the number of minimal sets times two. These are also the number of examples that the LSTM is evaluated on. We do not include inanimate-noun cases in our evaluations for now, since these are much more difficult to replicate cross-linguistically. Indeed, grammatical gender is a confounding variable which---according to preliminary experiments---does have an effect on model performance. Additionally, Hebrew has differing inflections depending on the combination of the subject and object noun genders, which means that we rarely have all needed inflections in the vocabulary.

We have differing numbers of examples per-language for similar cases. The reasoning for this is two-fold: (1) direct translations do not exist for all English items in the evaluation set of \citet{marvinlinzen18}, so we often must decide between multiple possibilities. In cases where there are two translations of a noun that could reasonably fit, we use both; if we have multiple possibilities for a given verb, we use only one---the most frequent of the possible translations. If no such translation exists for a given noun or verb, we pick a different word that is as close to the English token is possible in the same domain.

Reason (2) is that many of the nouns and verbs in the direct translation of the evaluation sets do not appear in the language models' vocabularies. Thus, some nouns or focus verbs would effectively be \texttt{<unk>}s if left in, rendering that particular example unusable. In such cases, if a given noun/verb is not the vocabulary, we pick a similar noun from the same domain if one exists; if a similar item does not exist in the vocabulary, we choose some common noun in that language's vocabulary that has not already been used in the evaluation set.

We use a similar process to add new verbs, but sometimes, third-person singular and plural inflections of similar verbs did not exist in the vocabulary. In such cases, we used a similar verb if possible (e.g., `dislike' would be reasonably similar in distribution and meaning to `hate'), but if no such similar verb exists in the vocabulary, we do not replace it. A similar process is used for closed classes like prepositions: if no sufficient replacement exists in the vocabulary, it is not replaced.

Table~\ref{tab:examples_bert} contains the number of examples used by BERT and mBERT to calculate examples. Important to note is that for these evaluations, we use stimulus types containing both animate \emph{and} inanimate nouns to better match \citet{bertgoldberg19}'s experimental setup; this is why we have more examples for English in this table than for the LSTM evaluations. Including or excluding inanimate nouns was found to make no significant difference in the final scores (for BERT or mBERT) regardless, since the performance of the model never diverges by more than 0.02 for animate vs.\ inanimate stimulus types.

The variation in the number of examples across languages is due to many of the focus verbs not being in mBERT's vocabulary. We see the lowest coverage in general for Hebrew and (surprisingly) French; this is likely due to Hebrew script being a rarer script in mBERT and due to many of French's most common tokens being split into subwords, respectively. Russian also has relatively low coverage, having $0$ in-vocabulary target verbs for long VP coordination. None of our languages except English had any target verbs for Within an Object Relative Clause.
\end{document}